\documentclass[10pt,journal,compsoc]{IEEEtran}
\ifCLASSOPTIONcompsoc
  \usepackage[nocompress]{cite}
\else
  \usepackage{cite}
\fi

\usepackage{url}
\usepackage{epsfig}
\usepackage{graphicx}

\usepackage{etoolbox}
\usepackage{bm}
\usepackage{amsthm}
\usepackage{algorithm}
\usepackage{algorithmic}
\usepackage{subfigure}
\usepackage{xspace}

\newcommand{\MyMapTemplatePrefixc}[4]{\expandafter#1\csname#3#4\endcsname{#2{#4}}} 
\forcsvlist{\MyMapTemplatePrefixc {\def} {\mathcal}{c}} {A,B,C,D,E,F,G,H,I,J,K,L,M,N,O,P,Q,R,S,T,U,V,W,X,Y,Z}  

\newcommand{\MyMapTemplatePrefixtb}[5]{\expandafter#1\csname#4#5\endcsname{#2{#3{#5}}}} 
\forcsvlist{\MyMapTemplatePrefixtb {\def} {\tilde}{\mathbf}{t}} {A,B,C,D,E,F,G,H,I,J,K,L,M,N,O,P,Q,R,S,T,U,V,W,X,Y,Z}  
\forcsvlist{\MyMapTemplatePrefixtb {\def} {\tilde}{\mathbf}{t}} {0,1,a,b,c,d,e,f,g,h,i,j,k,l,m,n,o,p,q,r,s,u,v,w,x,y,z}  

\newcommand{\MyMapTemplateNoPrefix}[3]{\expandafter#1\csname#3\endcsname{#2{#3}}}
\forcsvlist{\MyMapTemplateNoPrefix {\def} {\mathbf} } {0,1,a,b,c,d,e, f, g, h, i, j, k, l, m, n, o, p, q, r, u, v, w, x, y, z} 
\forcsvlist{\MyMapTemplateNoPrefix {\def} {\mathbf} } {A,B,C,D,E,F,G,H,I,J,K,L,M,N,O,P,Q,R,S,T,U,V,W,X,Y,Z}  

\def\etc{\emph{etc.}\@\xspace}
\def\ie{\emph{i.e.}\@\xspace}
\def\eg{\emph{e.g.}\@\xspace}

\usepackage{booktabs} 
\usepackage[table,dvipsnames]{xcolor}
\definecolor{rowblue}{RGB}{220,230,240}

\begin{document}
%
\title{WebVision Challenge: Visual Learning and Understanding With Web Data}


\author{
Wen~Li,
Limin~Wang,
Wei~Li,
Eirikur~Agustsson,
Jesse~Berent,
Abhinav~Gupta,
Rahul~Sukthankar,
and~Luc~Van~Gool
\IEEEcompsocitemizethanks{
\IEEEcompsocthanksitem Wen Li, L. Wang, E. Agustsson, and L. Van~Gool are with
the Computer Vision Laboratory, ETH Zurich, Switzerland. \protect\\
 E-mail: \{liwen, limin.wang, aeirikur, vangool\}@vision.ee.ethz.ch
\IEEEcompsocthanksitem Wei Li and J. Berent are with the Google Research, Zurich, Switzerland. \protect\\
E-mail: \{jesse.berent, lwthucs\}@gmail.com
\IEEEcompsocthanksitem A. Gupta is with the The Robotics Institute, Carnegie Mellon University, Pittsburgh, USA. \protect\\
E-mail: abhinavg@cs.cmu.edu
\IEEEcompsocthanksitem R. Sukthankar is with the Google Research, USA. \protect\\
E-mail: sukthankar@google.com
}}

\IEEEtitleabstractindextext{%
\begin{abstract}
We present the 2017 WebVision Challenge, a public image recognition challenge designed for deep learning based on web images without instance-level human annotation. Following the spirit of previous vision challenges, such as ILSVRC~\cite{ImageNet}, Places2~\cite{Places2} and PASCAL VOC~\cite{Pascal}, which have played critical roles in the development of computer vision by contributing to the community with large scale annotated data for model designing and standardized benchmarking, we contribute with this challenge a large scale web images dataset, and a public competition with a workshop co-located with CVPR 2017. The WebVision dataset contains more than $2.4$ million web images crawled from the Internet by using queries generated from the $1,000$ semantic concepts of the benchmark ILSVRC 2012 dataset. Meta information is also included. A validation set and test set containing human annotated images are also provided to facilitate algorithmic development. The 2017 WebVision challenge consists of two tracks, the image classification task on WebVision test set, and the transfer learning task on PASCAL VOC 2012 dataset. In this paper, we describe the details of data collection and annotation, highlight the characteristics of the dataset, and introduce the evaluation metrics.
\end{abstract}

}

\maketitle

\IEEEdisplaynontitleabstractindextext
\IEEEpeerreviewmaketitle

\IEEEraisesectionheading{\section{Introduction}\label{sec:introduction}}
\IEEEPARstart{T}{he} recent success of deep learning has shown that a deep architecture in conjunction with abundant quantities of labeled training data is the most promising approach for most vision tasks~\cite{krizhevsky2012imagenet,HeZRS16,SzegedyLJSRAEVR15,long2015fully,ren2015faster, DosovitskiyFIHH15,TranBFTP15,SimonyanZ14,WangXW0LTG16,DongLHT16}. However, annotating a large-scale dataset for training such deep neural networks is costly and time-consuming, even with the availability of scalable crowd-sourcing platforms like Amazon Mechanical Turk. As a result, there are relatively few public large-scale datasets (\eg, ImageNet~\cite{ImageNet} and Places2~\cite{Places2}) from which it is possible to learn generic visual representations from scratch.

Thus, it is unsurprising that there is a continued interest in developing novel deep learning systems trained on low-cost data, including unlabeled images/videos~\cite{Vincent10, Radford15}, self-supervised and semi-supervised approaches~\cite{Doersch15, Wang15, Agrawal15, Noroozi16}, and methods that exploit weak and noisy labels from auxiliary sources~\cite{Chen15, Joulin16, Owens16, Krause16}. In particular, there is promising recent work on using the web as a source of supervision for learning deep representations for a variety of important computer vision applications, including image annotation, object detection and fine-grained classification~\cite{ Chen15, Joulin16, Krause16}.

Learning from web data differs from purely supervised or unsupervised learning because images and videos on the web are naturally accompanied with abundant meta data (such as surrounding text, title, tags, \etc) that can provide weak supervision without the tedium or expense of crowd-sourced manual label. While the existing works~\cite{Vijayanarasimhan08,  Li14, Chen15, Joulin16, Krause16} have shown advantages of using web data in various applications, their tasks and methodologies differ from each other, making it hard to identify key issues and effective ways when utilizing web data. Moreover, their results were often obtained using much more images or categories, making it difficult to understand the capacity of noisy web images for learning visual recognition models when compared with the human-annotated datasets.

With this challenge, we aim at promoting the advance of learning state-of-the-art visual models directly from the web. We build a new web image database called {\emph{WebVision}},  which contains more than $2.4$ million of web images crawled from the Internet (about $1$ million from Google Image search, and $1.4$ million from Flickr)  by using queries generated from the same $1,000$ semantic concepts as the benchmark ILSVRC 2012 datast. Meta information along with those web images (\eg, title, description, tags, \etc) are also crawled. A validation set and a test set, each containing $50,000$ human annotated images, are also provided to facilitate algorithmic development. The dataset is now available at \url{http://vision.ee.ethz.ch/webvision}.

Based on this new dataset, we host a public competition on visual recognition by learning deep models from web images. We also organize a workshop at the IEEE Conference on Computer Vision and Pattern Recognition (CVPR) conference 2017, and call for researcher all over the world to meet and discuss the competition results and key research issues in learning from web data. More details and updates on the workshop can be found at \url{http://vision.ee.ethz.ch/webvision/workshop.html}.

\begin{figure*}[t]
\centering
\includegraphics[width=1.0\textwidth]{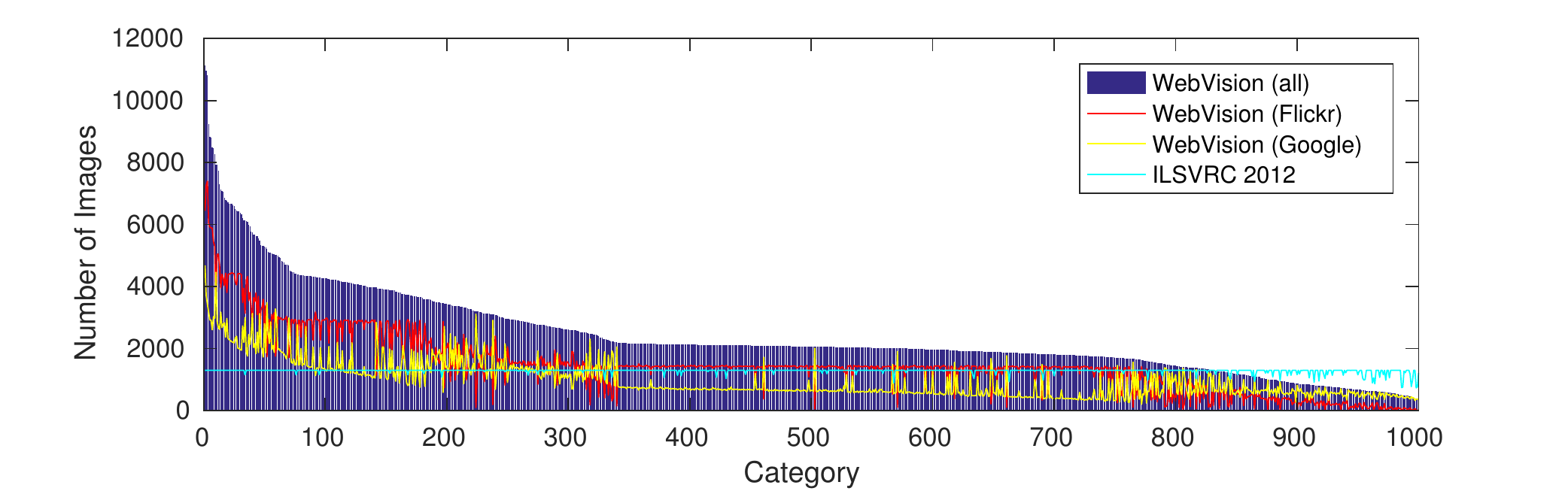}
\caption{Number of images per category of the WebVison dataset.}
\label{fig:hist}
\end{figure*}

\section{WebVision Dataset}
To study learning from web data, we build a large scale web image database called WebVision by crawling web images from the Internet. This new database is then used to investigate the potential of the web data for learning representations in this work. Next we will describe the details on the construction of the WebVison dataset, and then provide an analysis on it.

\subsection{Dataset Construction}
\textbf{Semantic Concepts:}
The first issue for building a new database is, what semantic concepts of web images shall we collect from the Internet to learn a generic representation? A successful example of labeled dataset is the ILSVRC 2012 dataset~\cite{ImageNet}, which consists of $1,000$ semantic concepts.  The representation learnt from those $1,000$ concepts of images exhibits good generalization ability, and it has been a common way to fine-tune CNN models learnt from ILSVRC 2012 dataset for various computer vision tasks, such as image classification~\cite{Oquab14}, object detection~\cite{GirshickDDM16}, object segmentation~\cite{ShelhamerLD17} and action recognition~\cite{SimonyanZ14}.  We construct our dataset by collecting web images from the same $1,000$ semantic concepts. Moreover, using the same $1,000$ semantic concepts as the ILSVRC 2012 dataset, it allows us to better understand the potential of the web data for learning representations by directly comparing with ones learnt from the ILSVRC 2012 dataset.

\textbf{Web Sources:} We consider two popular sources, the Google Image Search website\footnote{\url{http://images.google.com/}}, and the Flickr website\footnote{\url{http://www.flickr.com/}}. It has been shown in the literature that the images crawled from Google Image Search are effective for image categorization and representation learning~\cite{80MImages,Caltech256,Pascal,Places2,Chen15,Krause16}.

\textbf{Data Collection:}
We crawl web images from Flickr and Google Image Search based on queries generated from the $1,000$ synsets defined in the ILSVRC 2012 dataset~\cite{ImageNet}.  For the synsets containing multiple items, we treat each item as a query, and crawl images individually for each item in the synset of each category. Items with semantic ambiguity are revised or removed to avoid conflicts. For example, the synsets of ``n02012849'' and ``n03126707'' are the same, \ie, ``crane''. To eliminate the conflict, we augmented those two synsets as ``crane bird'', and ``crane truck, crane tower'', respectively.  Another example is ``loggerhead, loggerhead turtle, Caretta caretta", where ``loggerhead" may cause ambiguity (it also refers to a species of bird), and thus was removed. In total, we obtain $1,631$ queries from the synsets of $1,000$ semantic categories. Due to the difference in interpreting the queries, we used different connection words for some Flickr queries and Google queries. A complete list of the queries for both websites has been included in our released dataset.

For the Flickr website, we use its text based image search portal, and crawl up to $2,000$ images for each query. We remove images where the short side is less than $500$ pixels, and finally obtain $1.6$M images.

For the Google Image Search website, we crawl as many images as possible for each query, which usually results in $600$--$1,000$ images for each query. After removing the invalid links, we obtained in total $1.1$M images.

For each crawled image, its class label is decided by the synset that its corresponding query belongs to. For example, for the images crawled by using ``crane bird'', its synset ID is ``n02012849'', which has label $135$ using the ILSVRC label set. Since the image search results can be noisy, the training images may contain significant outliers, which is one of the important research issues when utilizing web data (see quantitative results in Section \ref{sec:dataset_anlaysis} and \ref{sec:experiments}).

\begin{figure}[t]
\centering
\subfigure[Flickr image]{
\includegraphics[width=0.22\textwidth,, height=80pt]{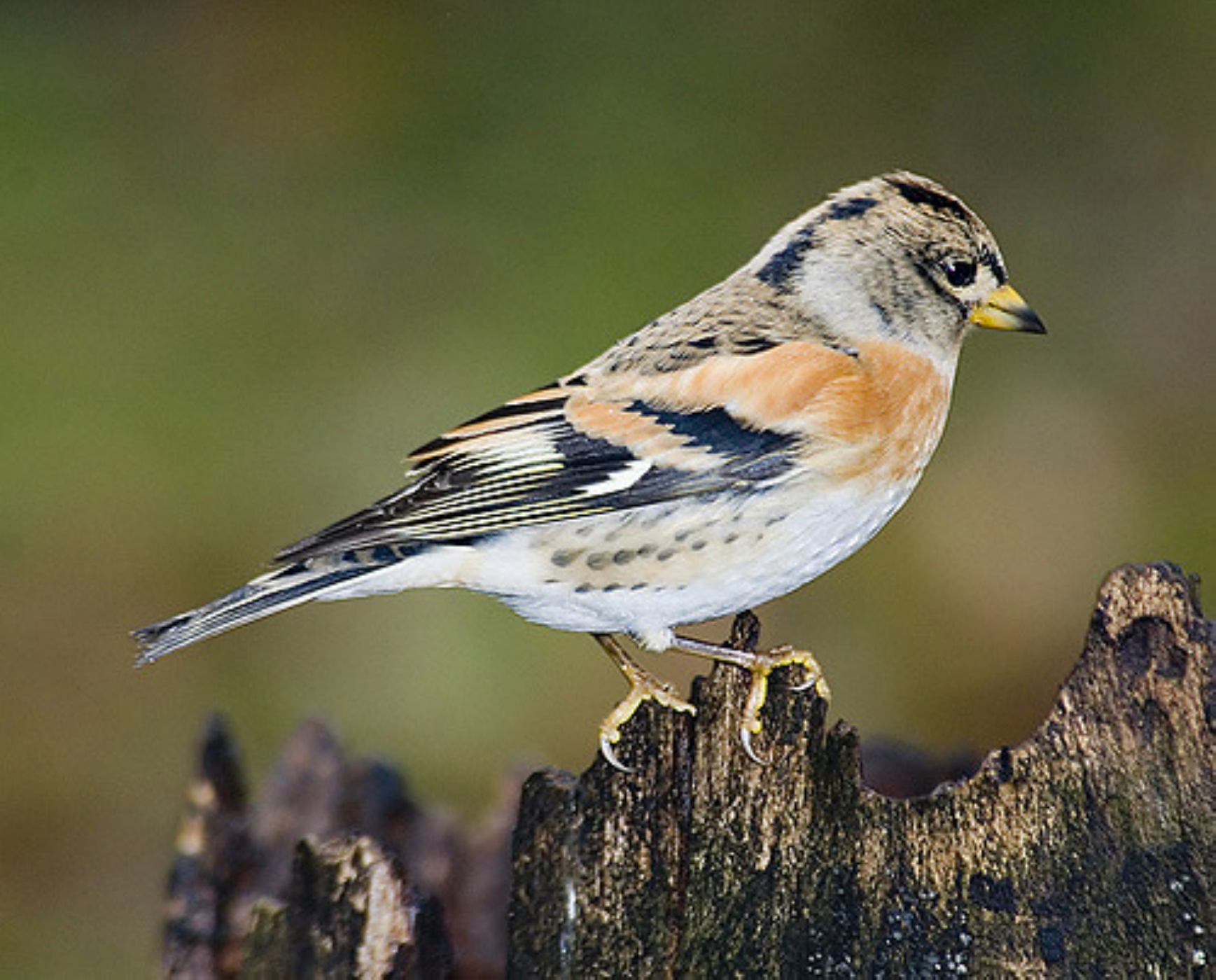}
}
\subfigure[Google image]{
\includegraphics[width=0.22\textwidth, height=80pt]{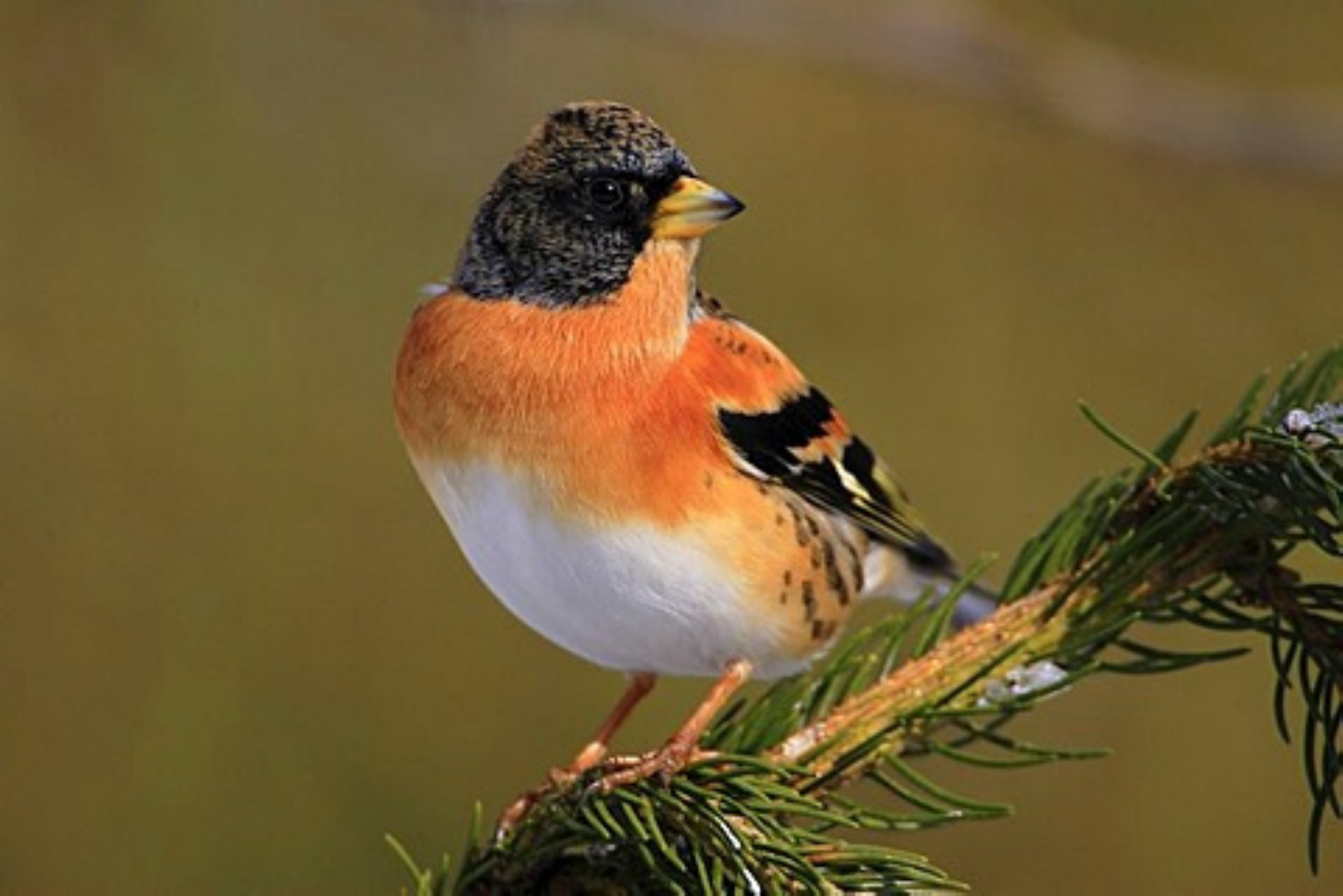}
}
\caption{Examples of image meta information from Flickr and Google. The meta-information associated with these two images is: (a) title: \emph{``Brambling''}; description: \emph{``Brambling - Fringilla montifringilla Russia, Moscow region, Saltykovka, 10/13/2007''}; tags: \emph{"Brambling", "Fringilla montifringilla"}; (b) title: \emph{``High Quality Stock Photos of brambling''}; description:\emph{``Brambling, male, North Rhine-Westphalia, Germany / (Fringilla montifringilla) /''}. }
\label{fig:meta}
\end{figure}

\textbf{Meta Information:} One advantage of web images is the abundant textual information, which usually contains valuable semantic information about the images, and has been shown to be quite useful for image categorization in the literature\cite{Schroff11,Li14,Joulin16}. For each Flickr image, we download its accomplished textual information, including \textit{title}, \textit{description}, \textit{tags}, \etc Geographical information and camera information is also included if it is available. For Google images, the \textit{title} and \textit{description} along with each image are crawled. An example of the meta information associated with images from both sources crawled using the query ``\emph{brambling}'' are shown in Figure~\ref{fig:meta}.

\textbf{Validation and Test Sets:} To facilitate algorithmic development, we also split a subset from the crawled images, and annotate a validation set and a test set. We randomly split out $200,000$ images ($200$ images per category), and put them along with their noisy labels on the Amazon Mechanical Turk (AMT) platform \footnote{\url{http://www.mturk.com/}}. The users are asked to verify if the label provided with each image is correct or not. Each image is annotated by three users, and is considered as an inlier image if more than two users agree. For concepts with less than $100$ inlier images, we continue to split a number of images from the crawled data, and send to AMT for annotation. Finally, we obtain in total $100,000$ human-annotated images, where each of the $1,000$ categories contains $100$ images. We then equally split it into two sets, a validation set and a test set, each containing $50,000$ images, \ie, $50$ images per category.

The remaining images are used as the training set. To ensure that there is no overlap between the training set and validation or testing set, we perform near-duplicate image detection and remove near duplicate images from the training set~\cite{Places2}. Finally, the training set of WebVision database contains in total 2,439,574 images, in which 1,459,125 images are from Flickr and 980,449 images are from Google Image Search.

\subsection{Dataset Analysis}
\label{sec:dataset_anlaysis}
\textbf{Category Distribution: } We plot the number of images per category for our WebVision database as well that for the ILSVRC 2012 dataset in Figure~\ref{fig:hist}. The number of images per category in the ILSVRC 2012 dataset is restricted no more than 1,300. For our WebVision database, and the number of images per category varies from $300$ to more than $10,000$. the number of images per category depends on both the number of queries generated from the synset for each category, and also the availability of images on Flickr and Google. Usually a category with many queries contains more images.

\begin{figure}[t]
\centering
\includegraphics[width=0.44\textwidth]{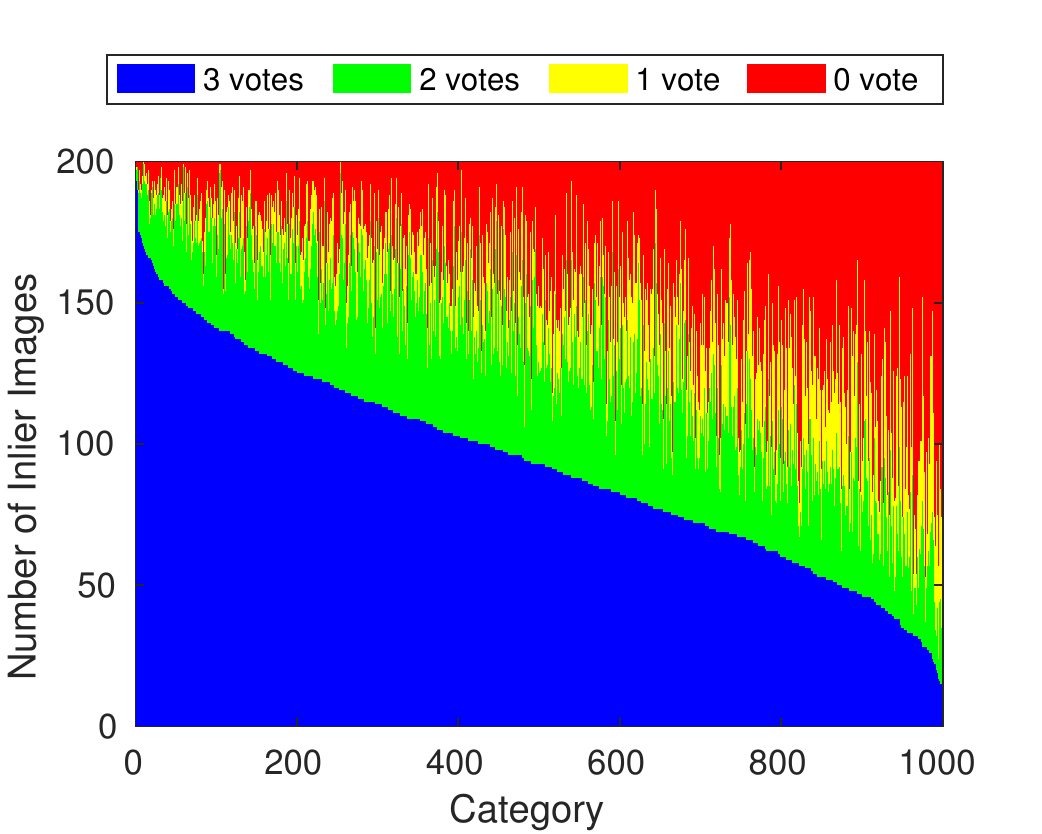}
\caption{Number of inlier images among 200 images per category of the WebVison dataset, sorted by number of ``3 votes" images in descend order. }
\label{fig:truth_num}
\end{figure}

\textbf{Domain Difference: } Examples of Flickr and Google images in our WebVision database can be found at our website~\url{http://vision.ee.ethz.ch/webvision}. Generally, the Google images are usually with a clean background, and the objects/targets in the image are captured with a clear shot. In contrast, the images from Flickr are usually captured with various backgrounds in the wild, and the objects/targets are sometimes with small sizes. As a comparison, the ILSVRC 2012 dataset is filtered with human annotation, so the objects/targets are usually clearly visible with diverse backgrounds. A quantitative analysis on the domain difference between WebVision and ILSVRC 2012 datasets can be found in Section~\ref{sec:experiments}.

\textbf{Noisy Labels:} To investigate how noisy the labels of web images are, we take the annotation results from the first round (200K images) as an example, and plot the user votes in Figure \ref{fig:truth_num}. Each vote indicates that a user agrees the provided label is correct, and images with more than 2 out of 3 votes are considered as true inlier images.

From the figure, we observe that the crawled web images contain a considerable amount of outliers. About 20\% of images are considered as true noisy images (\ie, ``0 vote''), and the  inlier images (\ie, ``3 votes'' and ``2 votes'') take only 66\% of the total images.
Moreover, the number of inlier images varies a lot in different categories.
The  cleanest category is ``867 -- Tractor'' which contains $199$ inlier images among $200$ split images. The worst one is ``627 -lighter, light, igniter, ignitor'', which has only $24$ inlier images.

\section{Tasks and Evaluations}\label{sec:experiments}
The goal of this challenge is to advance the area of learning knowledge and representation from web data. The web data not only contains huge numbers of visual images, but also rich meta information concerning these visual data, which could be exploited to learn good representations and models. We organize two tasks to evaluate the learned knowledge and representation: (1) WebVision Image Classification Task, and (2) Pascal VOC Transfer Learning Task. The second task is built upon the first task. Researchers can participate into only the first task, or both tasks.

\subsection{WebVision Image Classification Task}
The WebVision dataset is composed of training, validation, and test set. The training set is downloaded from Web without any human annotation. The validation and test set are human annotated, where the labels of validation data are provided but the labels of test data are withheld. To imitate the setting of learning from web data, the participants are required to learn their models solely on the training set and submit classification results on the test set. The validation set could only be used to evaluate the algorithms during development. Each submission will produce a list of $5$ labels in the descending order of confidence for each image. The recognition accuracy is evaluated based on the label which best matches the ground truth label for the image. Specifically, an algorithm will produce a label list: $c_i, i=1,.\ldots, 5$ for each image and the ground truth labels of the image are: $y_j, j=1,\ldots, n$ with $n$ class labels. The error of this prediction is defined as:
$$
E = \frac{1}{n}\sum_{j=1}^n\min_i d(c_i, y_j)
$$
The $d(c_i, y_j)$ is calculated as $0$ if $c_i=y_j$ and $1$ otherwise. The final errors of the algorithm is the average corresponding error across all test images. For this version of the challenge, there is only one ground truth label for each image (\ie, $n=1$).
\subsection{Pascal VOC Transfer Learning Task}
This task is designed for verify the knowledge and representation learned from the WebVision training set on the new task. Hence, participants are required to submit results to the first task and transfer only models learned in the first task. We choose the image classification task of Pascal VOC 2012 to test the transfer learning performance. Participants could exploit different ways to transfer the knowledge learned in the first task perform image classification Pascal VOC 2012. For example, treating the learned models as feature extractors and learning the SVM classifier based on the features. The evaluation protocol strictly follows the previous Pascal VOC, \ie, using the mean of average precision (mAP) as the evaluation metric (see \url{http://host.robots.ox.ac.uk/pascal/VOC/voc2012/htmldoc/devkit_doc.html#sec:ap}). 
\section{Experiments}\label{sec:experiments}
Details of experimental evaluation and in-depth analysis will be updated at the hompage of the WebVision dataset \url{http://vision.ee.ethz.ch/webvision}.

\section*{Acknowledgments}
This work is supported by the Computer Vision Laboratory at ETH Zurich, and the Google Research, Zurich. The authors gratefully thank NVIDIA Corporation for donating the GPUs used in this project.

\ifCLASSOPTIONcaptionsoff
  \newpage
\fi



%
\bibliographystyle{IEEEtran}
\bibliography{webvision_bib}

\end{document}